%% file: main.tex
\documentclass[runningheads]{llncs}

\usepackage{eccv} 

\usepackage{graphicx}
\usepackage[normalem]{ulem}
\usepackage{wrapfig}
\usepackage{multirow}

\usepackage[accsupp]{axessibility}  

\usepackage[breaklinks,colorlinks,citecolor=eccvblue]{hyperref}
\usepackage[T1]{fontenc}
\usepackage[english]{babel}
\usepackage{fixmetodonotes}
\usepackage{booktabs}
\usepackage{mathtools}
\usepackage{algorithm2e}
\usepackage[inline,shortlabels]{enumitem}
\usepackage{amsfonts}
\usepackage{stmaryrd}
\usepackage{booktabs}
\usepackage[dvipsnames]{xcolor}
\usepackage[english]{babel}
\usepackage{orcidlink}

\AtEndPreamble{\input{math_commands}}

\newcommand{\method}{ReGentS}

\begin{document}
\title{ReGentS: Real-World Safety-Critical Driving Scenario Generation Made Stable}

\titlerunning{ReGentS}

\author{Yuan Yin\inst{1}\orcidlink{0000-0003-1515-0696} \and
Pegah Khayatan\inst{1}\thanks{Work done during an internship at Valeo.ai.}\orcidlink{0009-0001-9707-4836} \and Éloi Zablocki \inst{1}\orcidlink{0000-0003-2757-2036} \and 
Alexandre Boulch\inst{1}\orcidlink{0000-0002-4196-9665} \and Matthieu Cord \inst{1,2}\orcidlink{0000-0002-0627-5844}}
\authorrunning{Y. Yin et al.}
\institute{Valeo.ai, Paris, France \\
\email{\{firstname.lastname\}@valeo.com} \and 
Sorbonne Université, CNRS, ISIR, F-75005 Paris, France}

\maketitle 

\begin{abstract}
Machine learning based autonomous driving systems often face challenges with safety-critical scenarios that are rare in real-world data, hindering
their large-scale deployment. While increasing real-world training data coverage could address this issue, it is costly and dangerous. This work explores generating safety-critical driving scenarios by modifying complex real-world regular scenarios through trajectory optimization. We propose \method{}, which stabilizes generated trajectories and introduces heuristics to avoid obvious collisions and optimization problems. Our approach addresses unrealistic diverging trajectories and unavoidable collision scenarios that are not useful for training robust planner. We also extend the scenario generation framework to handle real-world data with up to 32 agents. Additionally, by using a differentiable simulator, our approach simplifies gradient descent-based optimization involving a simulator, paving the way for future advancements. The code is available at \url{https://github.com/valeoai/ReGentS}.
\end{abstract}
\input{figs/teaser}

\section{Introduction}
\label{sec:intro}
With the rapid advancement of machine learning (ML), autonomous driving systems are becoming increasingly proficient, particularly through neural network-based methods \cite{chen2024endtoendautonomousdrivingchallenges}. 
However, their performance relies heavily on the extent of training data coverage, and they may struggle with unseen, safety-critical scenarios rarely encountered in real-world data. Improving out-of-distribution generalization remains a important topic within the ML community \cite{OOD-YangZLL2021}. This challenge hinders the large-scale deployment of autonomous driving systems, which must meet strict safety standards. One solution is to enhance real-world data coverage by collecting more diverse data, such as through event data recorders (EDRs) \cite{tesla}. However, collecting data specifically for safety-critical scenarios is dangerous, costly, and fraught with privacy concerns. Alternatively, generating these scenarios offers a solution. However, most existing approaches \cite{WangPTMS0RU21,huang2024versatilesceneconsistenttrafficscenario,xu2023diffscene,RempePGFL22,hanselmann2022king} have been benchmarked on synthetic data, which features scenarios of limited complexity and often involves few vehicles. Additionally, the generation process is often unconstrained, which may produce unrealistic solutions as shown in \cref{fig:teaser-king} or scenarios not useful for enhancing ML-based planner's robustness.

In this work, we introduce \method{}, an approach for generating more stable safety-critical driving scenarios based on real-world data. We study some common choices in existing approaches, such as the cost that induces collisions. By focusing on KING \cite{hanselmann2022king} and refactoring it at a larger scale and in a more realistic setting with real-world data based simulator, Waymax \cite{waymax}, we identify following areas for improvement: \begin{enumerate*}[(1)]
    \item Unrealistic swinging trajectories may be produced, and
    \item Many generated collisions involve the ego vehicle being rear-ended, which cannot in practice be avoided and thus does not provide useful corner cases for planner training.
\end{enumerate*}
We address these issues by analyzing the optimization process of \cite{hanselmann2022king} and adding constraints to produce more stable trajectories. Additionally, we introduce a heuristic to avoid obvious collision scenarios. Our approach enhances stability of generated trajectories with more realistic driving behavior compared to the unconstrained method, as shown in \cref{fig:teaser-king,fig:teaser-ours}.

In a technical viewpoint, we implement and simplify the setup using a single differentiable simulator \cite{waymax}. This allows for direct optimization through gradient descent, making the process simpler, unified, and easier to extend. The differentiable simulator also avoids the heavy interaction between a differentiable proxy simulator and non-differentiable one like CARLA \cite{Dosovitskiy17}, and eliminating inconsistencies that may arise between the two simulators.

In summary, our contributions are:

\begin{itemize}
\item We propose solutions to two main issues identified in existing approaches: solution stability and the choice of adversarial vehicles. Specifically, we adapt the optimization process to enhance stability with tailored analysis and introduce heuristics based on the position of the background vehicle.
\item We implement a trajectory optimization-based scenario generation pipeline on a large-scale, differentiable simulator for real scenarios and data, facilitating its use in further studies.
\item We demonstrate that \method{} selects better adversaries and generates more stable trajectories in certain cases. Quantitatively, it produces more safety-critical scenarios, which is useful for 
fine-tuning ML-based planners.
\end{itemize}

The paper is organized as follows: \cref{sec:related} presents previous and related literature, \cref{sec:setting} describes the addressed problem, \cref{sec:methodo} is a complete description of \method{} and \cref{sec:exp} is dedicated to experimental evaluation of our approach.

\section{Related Work}
\label{sec:related}

The literature has been concentrating on trajectory generation to create ego-adversary collisions, which we categorize into the following approaches. 

\paragraph{Black-box Optimization Approaches.} Early methods treat the problem as a black-box adversarial attack. For example, AdvSim \cite{WangPTMS0RU21} uses Bayesian optimization. It employs a trajectory sampling method that rejects non-physically realizable trajectories, and proposes a cost function to bias the sampling to create collisions.
    
\paragraph{Generative Model-based Approaches.} Generative models learn distributions based on provided data. In recent advances of complete scenario generation, such as \cite{Pronovost2023}, the results may lack diversity and physical feasibility. More specifically, in safety-critical case generation, these approaches typically involve two steps: (1) training a generative model to sample from a regular distribution, and (2) biasing the generation to create collisions using hand-crafted losses. STRIVE \cite{RempePGFL22} uses a conditional variational autoencoder (CVAE) to learn a latent space of regular scenarios, which can then be adjusted to generate collisions. CAT \cite{pmlr-v229-zhang23g} introduces a resampling strategy to bias generative models towards collision scenarios. Recent efforts, such as \cite{xu2023diffscene,huang2024versatilesceneconsistenttrafficscenario}, have introduced diffusion models that generate trajectories by diffusing states/actions from Gaussian noise. These models compute the gradient of a collision-inducing cost \wrt the states/actions to guide the diffusion process using classifier guidance.
    
\paragraph{Model-Based Trajectory Optimization Approaches.} Unlike previous approaches, model-based methods such as KING \cite{hanselmann2022king} optimize trajectories directly by leveraging a kinematic model and actions.

The costs used by all the approaches above are derived from the collision condition \cref{eq:collision}. 
They are mainly benchmarked on simulators with limited realism, such as CARLA \cite{Dosovitskiy17}. Instead of injecting bias into a distribution as in (a) and (b), we choose to modify existing scenarios directly, as in (c). Unlike the unconstrained optimization in \cite{hanselmann2022king}, we impose additional constraints.

\section{Problem Setting} 
\label{sec:setting}
In this section, to put important elements in order, we provide a formal definition of safety-critical scenario generation common in the domain. In the following, without loss of generality, we focus on the bird's-eye-view (BEV) scenario.

\subsection{Scenario Definition}
We define a driving scenario $s \in \mathcal{S}$ as being composed of time-invariant contexts and time-variant series. 

\textit{Time-invariant contexts} include essential information and parameters for constructing the scenario background. This includes the road graph $g$, which specifies the positions of road features such as boundaries and lane center lines. Additionally, vehicle (agent) metadata $\{v^{(i)}\}_{i \in \llbracket 0, n \rrbracket}$ is provided, where $i$ indicates the $i$-th agent in the scenario. The metadata $v^{(i)}$ includes the width and length of agent's bounding box. The agent designated as $i=0$ is the ego agent, while agents $i \in \llbracket 1, n \rrbracket$ represent the $n$ background (potentially adversarial) agents. 

\textit{Time-variant series} correspond to two categories:
traffic signals and trajectories. Each series is uniformly discretized over $T$ time steps $k\in\llbracket0, T-1\rrbracket$ with interval $\delta t$, meaning that $k$ represent $k\delta t$. 
Traffic signals are a series $w = (w_{k})_{k \in \llbracket 0, T-1 \rrbracket}$, including, \eg, traffic light statuses. The trajectories of the ego and background agents are denoted by $\{q^{(0)}_\pi\} \cup \{q^{(i)}\}_{i \in \llbracket 1, n \rrbracket}$, where each $q^{(i)}$ is represented as a series of states, \ie, $q^{(i)} = (q^{(i)}_{k})_{k \in \llbracket 0, T-1 \rrbracket}$. The ego trajectory $q^0_\pi$ is unrolled with a planner $\pi$. Typically, each state $q_k$ includes the position $x_k$, the speed $u_k$, and the orientation angle $\psi_k$, \ie, $q_k = (x_k, u_k, \psi_k) \in \gQ = \R^2 \times \mathbb{R} \times [-\pi, \pi]$. To summarize, a scenario is defined as $s = (g, \{v^{(i)}\}, w, \{q^{(0)}_\pi\} \cup \{q^{(i)}\}_{i > 0})$. By abuse of notation, we denote the scene at a time step $k$ by $s_k = (g, \{v^{(i)}\},  w_k, \{q^{(0)}_{k,\pi}\} \cup \{q_k^{(i)}\}_{i > 0})$.

\subsection{Collision-Style Safety-Critical Condition} 

A widely accepted criterion of a ``safety-critical'' scenario in the field (see \cref{sec:related} for examples) is that there should be a collision between the ego agent $i=0$ and any background agent $i>0$. For a scenario $s$, it is commonly defined based on the distance between the bounding boxes of agents $i$ and $j$ at time step $k$: \begin{equation}
        d^s_\text{BB}(i, j, k) = d\left(\operatorname{BB}(q^{(i)}_{k, \pi}, v^{(0)}), \operatorname{BB}(q^{(j)}_{k}, v^{(i)})\right)
\end{equation}
where $\operatorname{BB}$ computes the agent's bounding box from the state $q_k$ (position and orientation) and the agent's metadata $v^{(i)}$, and the distance $d$ is the Euclidean distance in $\mathbb{R}^2$ between closest points on the bounding polygons. Therefore, a scenario is considered safety-critical if
\begin{equation}
\label{eq:collision}
    \exists k\in \llbracket0,T-1\rrbracket, i>0, d^s_\text{BB}(0, i, k) \leq 0.
\end{equation}
indicating an overlap between the ego agent $i=0$ and any adversary $i>0$ at some time step $k$.

\subsection{Safety-Critical Scenario Generation with Regular Data}

To generate scenario with collisions involving the ego agent, we use a dataset of scenarios $\gD = \{s\}_{s \sim p_{\gS}(s)}$, collected from the unknown real-world scenario distribution $p_{\gS}$. These data are often retrieved using well-equipped vehicles driven by trained drivers. Publicly available datasets typically contain common scenarios under regular driving conditions, often lacking safety-critical situations.

However, generating the entire scenario $s$ is difficult for safety-critical situations due to the lack of safety-critical reference data. In practice, scenario generation is further narrowed down to the conditional generation of trajectories with ego-adversary collisions. This involves generating adversary trajectories that are both plausible and satisfy the collision condition in \Cref{eq:collision}, given the road graph $g$, agent metadata $v$, and potentially the traffic light statuses $w$.

\section{Methodology}
\label{sec:methodo}

In this section, we present \method{} as follows: We begin by introducing the model assumptions in \cref{sec:assumptions}. Next, we provide an overview of \cite{hanselmann2022king}, the prerequisite of our work in \cref{sec:king}. Finally, we describe our \method{}, which results from our analysis of the optimization process, detailed in \cref{sec:analysis}.

\begin{figure}[t]
    \centering
    \includegraphics[width=0.7\linewidth]{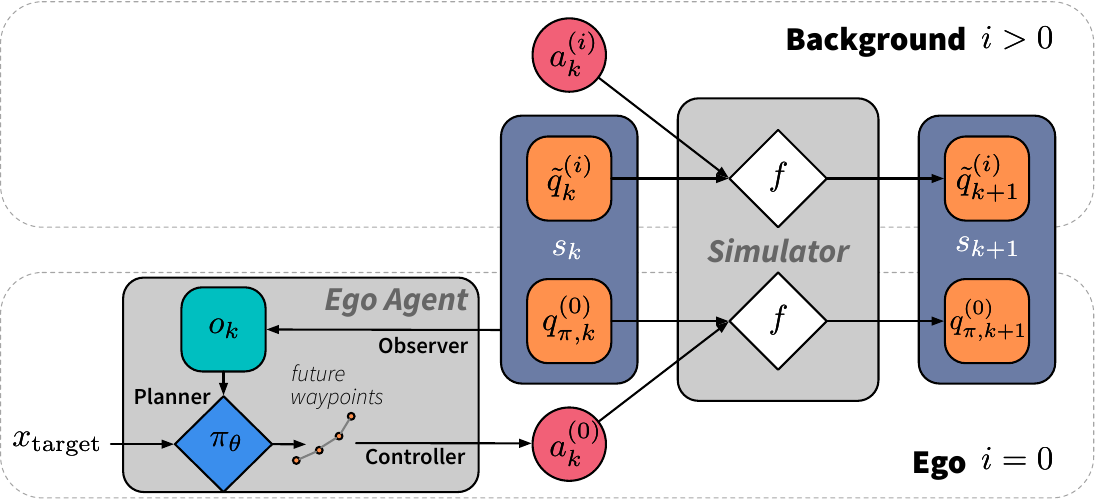}
    \caption{{\textbf{Forward Model Assumption of \method{} Through a Differentiable Simulator}}, accompanied by the agent's structure for state transition from $s_k$ to $s_{k+1}$. The ego agent ($i=0$) generates an observation $o_k$ by rasterizing the scenario onto a uniform grid. The planner uses this observation and a target point $x_\text{target}$ to predict the ego agent's action $a_k^{(0)}$ with a controller, which is fed into the kinematic model $f$. For background agents ($i>0$), the model $f$ estimates their actions $a^{(i)}$ and reconstructs their logged trajectory $\tilde{q}^{(i)}$. Arrows represent the computational graph which enables the backward autodiff.}
    \label{fig:forward-model}
\end{figure}

\subsection{Model Assumption} \label{sec:assumptions} The trajectory of each background agent ($i>0$) is supposed to be controlled by a series of actions through a discretized kinematic dynamics model:
\(
    q^{(i)}_{k+1} = q^{(i)}_{k} +  f(q^{(i)}_k, a^{(i)}_k)\delta t
\)
where $f$ is the vehicle's dynamics model (\eg, a Bicycle Model considered in \cite{hanselmann2022king}), $a^{(i)} = (a^{(i)}_{k})_{k\in\llbracket0, T-2\rrbracket}$ is a series of $T-1$ actions. We denote this action-conditioned trajectory as $\tilde{q}^{(i)} = (q^{(i)}_0) \mathbin\Vert (\tilde{q}^{(i)}_{k})_{k \in \llbracket 1, T-1 \rrbracket}$ where $\tilde{q}^{(i)}_ {k} = q^{(i)}_0 + \sum_{s=0}^{k-1} f(q_s^{(i)}, a^{(i)}_s)\delta t$. The action-conditioned trajectory $\tilde{q}^{(i)}$ should recover the original one $q^{(i)}$, \ie, $\tilde{q}^{(i)}\approx q^{(i)}$. A scenario $s$ is therefore extended as $s = (g, w, \allowbreak \{v^{(i)}\}, \{q^{(0)}_\pi\} \cup \{\tilde{q}^{(i)}\}_{i>0}, \{a^{(i)}\}_{i>0})$. 
Note that the action series $a^{(i)}$ are typically not available in most datasets and require to be estimated given the kinematic model $f$ and original trajectories. The ego agent is assumed as a mapping $\pi\colon (s_k, x_\text{target}) \mapsto a_k$, which can be further decomposed into an observer-planner-controller structure (see details in \cref{sec:implem}). We depict this forward scheme in \cref{fig:forward-model}.

\subsection{Prerequisite: KING}
\label{sec:king}

Our method extends KING \cite{hanselmann2022king}, which is based on the setting described in \cref{sec:setting} and generates new scenarios by modifying the trajectories of background agents in each $s \in \gD$ using a kinematic model and estimated agent actions. The core of this approach is detailed below.

\paragraph{Scenario Generation as Optimization Problem.} The scenario generation problem, as defined in KING, involves solving an optimization problem using a cost function $\gC(s)$ for a given scenario $s$ that encourages background agents to provoke a collision with the ego agent in an initially collision-free scenario. The optimization problem is formulated as:
\begin{equation}
\label{eq:optim}
    a^* = \argmin_a \gC(s)  \text{\: \sut \: } \forall i\in\llbracket0,n\rrbracket, k\in\llbracket0, T-2\rrbracket, q_{k+1}^{(i)} = q_{k}^{(i)} +  f(q_k^{(i)}, a_k^{(i)})\delta t
\end{equation}
where $a$ represents the action series of all background agents $\{a^{(i)}\}_{i>0}$. 
Solving this problem finds the actions that minimize the cost, subject to the physical dynamics model constraints. 

\paragraph{Cost Definition.} \cite{hanselmann2022king} decomposes the cost $\gC$ into two parts:
\begin{enumerate*}[(a)]
    \item \textit{Collision Induction}: A cost $\gC^\textit{ego}_\textit{col}$ is introduced to encourage collision between the ego agent and one of the background agents;
    \item \textit{Background Agent Regularization}: Two costs  $\gC^\textit{adv}_\textit{col}, \gC^\textit{adv}_\textit{dev}$ are introduced to avoid collisions among background agents and prevent the background agents from deviating from the drivable area. 
\end{enumerate*}

The cost (a) considers the average distance over all time steps $k\in\llbracket0, T-1\rrbracket$ between the ego agent ($i=0$) and each background agent ($i>0$) to indicate how likely a collision can be provoked:
\begin{equation}
\label{eq:ego_col}
\gC^\textit{ego}_\textit{col}(s) = \min_{i\in\llbracket 1, n\rrbracket} \frac{1}{T}\sum_{k=0}^{T-1} d^{s}_\text{BB}(0, i, k).
\end{equation}
This function selects the closest background agent to minimize its distance to the ego. 
Regularization (b) is intended to avoid unwanted solutions. This first term discourages collisions between background agents using distance-based repulsion:
\(\gC^\textit{adv}_\textit{col}(s) = -\min_{i, j\in\llbracket1, n\rrbracket, k\in \llbracket0, T-1\rrbracket} \allowbreak \min(\tau, d^s_\text{BB}(i, j, k))\), which measures the minimum distance between any pair of background agents over time, lower-bounded by threshold $\tau$. The second term is intended to help background agents stay within the drivable area:
\(
     \gC^\textit{adv}_\textit{dev}(s) = \frac{1}{T} \sum_{k=0}^{T-1} \sum_{i=1}^n\sum_{c\in\{c_l\}_{l=1}^4} (m_{oob} * \gK)\big(\operatorname{BB}\allowbreak\big(q^{(i)}_k, v^{(i)}_k\big)[c]\big)
\). Here, $\gK$ is a 2D Gaussian kernel, and $\{c_l\}_{l=1}^4$ represents the four corners of the agent's bounding box. The kernel is displaced by the coordinates of each corner and convolved with the binary drivable area map $m_{oob}$. A higher cost indicates a greater distance of background agents' bounding boxes from the drivable zone.

\paragraph{Solving Optimization Problem Using Gradient Descent.} The optimization problem \Cref{eq:optim} is solved with gradient descent methods. The gradient of the cost function $\nabla_{a}\gC(s)$ is computed \wrt all action series $\{a^{(i)}\}_{i>0}$ across all time steps, which is feasible if the kinematic model is differentiable. The optimization iteration is stopped once the condition \Cref{eq:collision} is satisfied. 

\subsection{\method{}}

\begin{figure}[t]
    \centering
\subcaptionbox{Minimum Trap: The optimization fixates to the nearest static adversary. \label{fig:trap}}{\includegraphics[trim={0.648cm 0.654cm 0 0},clip,width=0.24\linewidth]{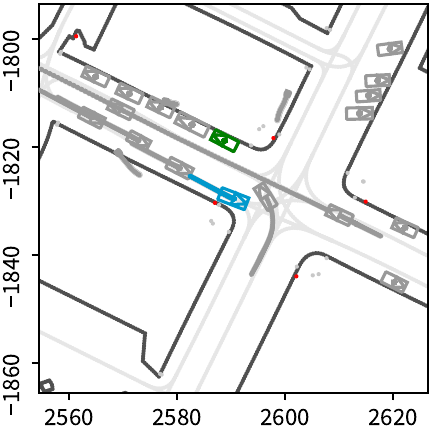}}
\subcaptionbox{Time-Averaged Distance Bias: Distance minimized at every time step without emphasis.\label{fig:time-averaged}}{\includegraphics[trim={0.648cm 0.654cm 0 0},clip,width=0.24\linewidth]{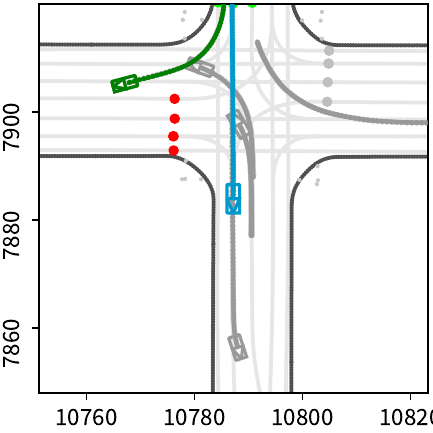}}
\subcaptionbox{Diverging Front Adversary: The distance minimization creates a trajectory that diverges from the ego's path.\label{fig:divergent-traj}}{\includegraphics[trim={0.648cm 0.654cm 0 0},clip,width=0.24\linewidth]{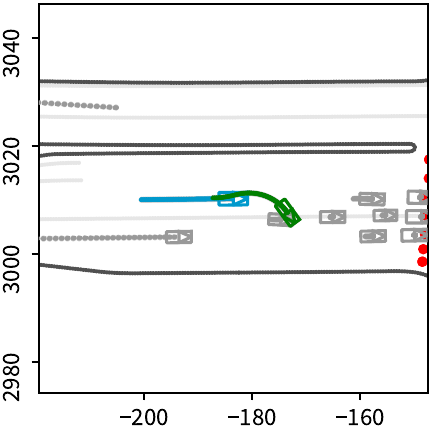}}
\subcaptionbox{Converging Rear-End Adversary: The adversary is on a collision course to rear-end the ego agent.\label{fig:rear-end}}{\includegraphics[trim={0.648cm 0.654cm 0 0},clip,width=0.24\linewidth]{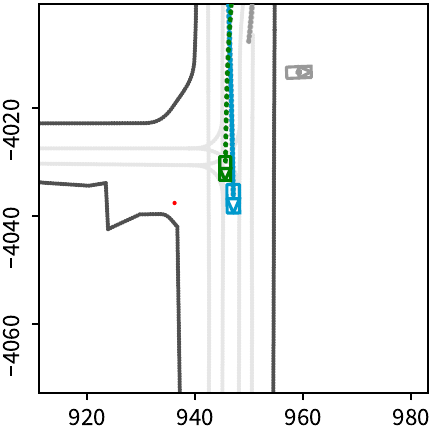}}

    \caption{\textbf{Illustration of Representative Issues with Unconstrained KING Method \cite{hanselmann2022king}.} The final state of each scenario is shown along with the trajectory history in dotted line. The ego agent (\textcolor{cyan}{blue}) is at the center, and the adversary agent (\textcolor{ForestGreen}{green}) represents the last optimized adversary.
    }
    \label{fig:example-with-issues}
\end{figure}
\subsubsection{Analysis.}
\label{sec:analysis}
We focus on studying the main collision induction cost \Cref{eq:ego_col} and its optimization. We first provide an analysis on the optimization problem, the optimization algorithm and the kinematic gradient of the main cost function.

\paragraph{Optimization Bias.}
The current approaches to safety-critical scenario generation are based on a distance-based loss derived from the acceptance condition of \Cref{eq:collision}. However, solving the optimization problem with the cost function \Cref{eq:ego_col} introduces some undesired biases:
\begin{itemize}
    \item \textit{Minimum Trap}: Although \Cref{eq:ego_col} does not explicitly designate an adversary, the hard minimum operator can cause the optimization to fixate on the adversary chosen in the first iteration. As shown in \cref{fig:trap}, the collision cost fixates on a stopped agent, while the agent turning at the intersection could have created a more interesting and avoidable collision.
    \item \textit{Time-Averaged Distance Bias}: As shown in \cref{fig:time-averaged}, the green adversary intended to turn right is pulled back towards the ego by the minimization of the time-averaged distance. A simpler solution would be to optimize only the trajectory before the adversary arriving at the intersection. This showcases that minimizing distance over all time steps may lead to ineffective solutions. 
    \item \textit{Limited Diversity in Adversary Trajectory}: Optimizing the action series $(a_k)_k$ of the chosen adversary relies solely on the optimization algorithm and its hyperparameters (e.g., learning rate or Adam's decay rates), limiting the diversity of results.
\end{itemize}

\begin{figure}[t]
    \centering
    \subcaptionbox{\label{fig:schema-throt} Negative gradient of distance cost \wrt throttle}{\includegraphics[height=3.65cm,trim={0.3cm 0 23.9cm 0},clip]{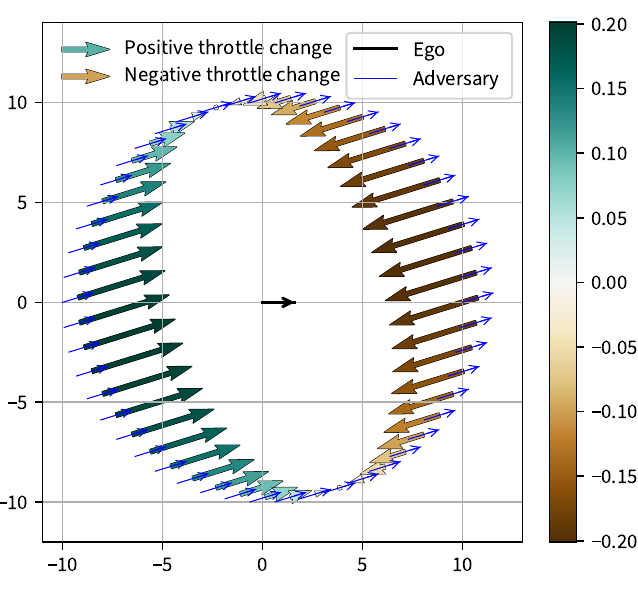}}\hfill
    \subcaptionbox{\label{fig:schema-steer} Negative gradient of distance cost \wrt steer}{\includegraphics[height=3.65cm,trim={12.2cm 0 10.8cm 0},clip]{imgs/divergence.pdf}}\hfill
    \subcaptionbox{\label{fig:schema-steer-after} Negative gradient of distance cost \wrt steer after one action update with the gradient shown in \subref{fig:schema-throt} and \subref{fig:schema-steer}. }{\includegraphics[height=3.65cm,trim={24.5cm 0 0.2cm 0},clip]{imgs/divergence.pdf}}
    \caption{\textbf{Negative Gradient of the Distance Between the Ego Agent ($\rightarrow$) and an Adversary (\textcolor{blue}{$\rightarrow$}) \wrt the Adversary's Actions.} In \subref{fig:schema-steer}, two interesting zones are highlighted. \textbf{Diverging Front Adversary} zone (\textcolor{red}{red}): Adversaries could pursue a diverging trajectory \wrt the ego agent. The negative gradient points to the diverging direction. The angle this zone is determined by the yaw of the adversary. \textbf{Converging Rear-end Adversary} zone (\textcolor{orange}{orange}): Adversaries are likely to rear-end the ego agent. After one gradient descent update from \subref{fig:schema-steer} to \subref{fig:schema-steer-after}, the negative gradient in the zoomed area intensifies, favoring the divergence of the adversary.
    }
    \label{fig:schema}
\end{figure}

\paragraph{Kinematic Gradient Bias.}
\label{sec:kinematic-gradient-bias}
Defining a distance-based loss to create collisions is convenient, but it can also lead to unrealistic behaviors with methods rely on kinematic gradient information \wrt agent actions, due to constraints on the degrees of freedom (DoF) of actions. We illustrate some issues with KING \cite{hanselmann2022king}, which relies heavily on kinematic gradients. Note that these issues may also affect recent classifier-conditioned diffusion models that generate actions to produce trajectories with a kinematic model \cite{xu2023diffscene,huang2024versatilesceneconsistenttrafficscenario}, using the same gradient to guide the generative model to induce collisions.

To illustrate the issue in the kinematic gradient, we present a simple scenario involving only the ego agent and a single adversary agent to be optimized. The adversary is shown at different locations but with the same distance to the ego, as shown in \cref{fig:schema}. Both agents are moving at a constant speed to the right. The key difference is their yaw: the ego's yaw is aligned with the $x$-axis, while the adversary's yaw is slightly offset. This scenario is particularly interesting when using real-world data because even if background agents are heading in the same direction as the ego, their yaws cannot be perfectly aligned due to noise in the data. The yaw offset shown in \cref{fig:schema} is chosen for better illustration and visualization.

In \cref{fig:schema}, we show the gradient descent directions (\ie, negative gradient) for the acceleration (\cref{fig:schema-throt}) and steering (\cref{fig:schema-steer}) of the adversary calculated from the distance between the ego and the adversary. As shown in \cref{fig:schema-throt}, the negative gradient of acceleration aligns well with the intuition of a typical driver: the adversary diminish its acceleration if it is ahead of the ego; it accelerates more if it is behind. However, the steering negative gradient, shown in \cref{fig:schema-steer}, is more complex. We focus on two specific zones for the adversary (red and orange), where undesired situations may occur. The angle of these zones corresponds to the difference in orientations between the ego and the adversary.
\begin{itemize}
    \item \textit{Diverging Front Adversary}: As illustrated in \cref{fig:divergent-traj}, the agent could have created a collision with the ego by just decelerating, while it did unnecessary turns which induced its collision with another background agent. The reason why this happens is when an adversary enters the red zone in \cref{fig:schema-steer} its negative gradient suggests it should turn away from the ego, causing the adversary's trajectory to diverge from the ego's. Subsequently, updating the actions with this negative gradient (see \cref{fig:schema-steer-after}) amplifies the yaw difference for agents already in the red zone, creating larger negative gradients which could push them further onto a more diverging trajectory (see the zoomed area in \cref{fig:schema-steer,fig:schema-steer-after}).
    As a result, agents in this zone are prone to showing unrealistic divergent behaviors, such as swinging turns between left and right. 
    \item \textit{Converging Rear-End Adversary}: In real-world driving environments, rear-end collisions are less relevant for improving the robustness of the planner. For example, if the ego vehicle is in dense traffic, it is almost impossible to avoid such a collision. However, optimizing adversaries in some areas is doomed to create a rear-end collision. As shown in \cref{fig:rear-end}, when an adversary is behind the ego, \ie, in the orange zone in \cref{fig:schema-steer}, it will be attracted to align with the yaw of the ego. It is very unlikely that the adversary in this area can overtake the ego because the tendency to diverge from the ego's past trajectory will be counteracted. Therefore, the optimized adversary is very likely to rear-end the ego, leading to uninteresting cases.
\end{itemize}

\subsubsection{\method{}' Remedy.}

Inspired by the previous analysis, we propose \method{}, which consists of several constraint rules designed to address some of the identified issues. Our goal is to make these criteria as flexible as possible.

\begin{wrapfigure}[13]{r}{0.35\linewidth}
    \centering
    \vspace{0em}
    \includegraphics[width=0.65\linewidth,trim={1cm 4cm 2cm 0},clip]{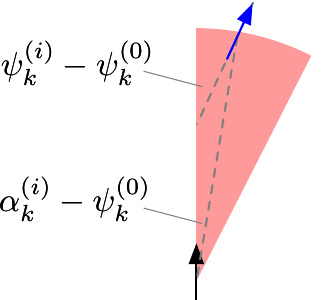}
    \caption{Illustration of how to determine if an adversary is in the red zone, with the red zone calculation shown. The ego agent is indicated by ($\rightarrow$) and the adversary by (\textcolor{blue}{$\rightarrow$}).}
    \label{fig:red-zone}
\end{wrapfigure}

\paragraph{Stop Front Divergence.} As discussed in \cref{sec:kinematic-gradient-bias}, front divergence may be caused by an aggrieving steer negative gradient favoring diverging trajectories. To mitigate this, we establish the following rule: we cancel the update for the steering action if an adversary stays (i) \textit{ahead of the ego agent} and (ii) its \textit{yaw is offset to the same side as its position relative to the ego} during a certain percentage of time steps, defined by a threshold $\tau_{\text{front}}$. 

The rule is illustrated in \cref{fig:red-zone}, featuring the same red zone shown in \cref{fig:schema}. At each time step $k$, the angle of the red zone is calculated by $\psi_k^{(i)} - \psi_k^{(0)}$. The deviation of the adversary's angular position relative to the ego's orientation, denoted by $\alpha_k^{(i)} - \psi_k^{(0)}$, is compared to the red zone angle to determine if the adversary is within the red zone at time $k$. This check is performed only for background agents with a yaw difference $\psi_k^{(i)} - \psi_k^{(0)} \in(-\frac{\pi}{8},\frac{\pi}{8})$ and whose angular position is $\alpha_k^{(i)} - \psi_k^{(0)}\in(-\frac{\pi}{8},\frac{\pi}{8})$ in front of the ego agent. 

\paragraph{Exclude Rear-End and Static Adversaries.} 
As explained in \cref{sec:analysis}, we may want to avoid rear-end collisions, which are unavoidable for the ego vehicle in some cases. In \method{}, we consider collisions in the orange zone as non-actionable and choose to exclude adversaries situated in this zone if they remain there at most time steps in the original scenario. The central angle of the disk sector zone is adjusted empirically as a hyperparameter. In our experiments, we set this zone as large as 45 degrees, with its axis of symmetry aligned with the 
$x$-axis. 

To avoid the problem where the optimization stick to a static adversary, we simply exclude all the static adversaries in the scenario from candidate to the optimization.  

\section{Experimental Results}
\label{sec:exp}

\subsection{Implementation for Real-World Scenarios} 
\label{sec:implem}

\paragraph{Data and Simulator.}
In our work, we uses the Waymo Open Motion Dataset (WOMD) \cite{womd} and Waymax \cite{waymax}, a differentiable simulator based on the scenario data in WOMD. The dataset comprises the scenarios recorded in cities, and we limit the maximum number of agents in a scenario to 32.
The main advantage of Waymax is its implementation with JAX \cite{jax2018github}, which is inherently differentiable, in contrast to CARLA, eliminating the need for another differentiable proxy simulator. In comparison, \cite{hanselmann2022king} uses a synthetic dataset featuring up to four background agents alongside the non-differentiable CARLA simulator. This setup significantly limited the flexibility of gradient calculations necessary to optimize trajectories through actions, thus necessitating a differentiable proxy simulator as a workaround.

\paragraph{Neural Network-Based Ego Agent.}

We employ a neural network-based ego agent tailored for WOMD. The agent $\pi\colon (s_k, x_\text{target}) \mapsto a_k$ determines actions based on the current scene $s_k$ at time step $k$ and a target point $x_\text{target}$. We adapt the AIM-BEV agent from KING, consisting of the following components:
\begin{enumerate*}[(1)]
    \item \textbf{Observer}: An observer $\operatorname{Obs}\colon s_k \mapsto o_k$ is implemented for WOMD to provide a rasterized ego-centered BEV observation $o_k$, oriented to ego's yaw direction.
    \item \textbf{Planner}: Using MobileNet-V3 \cite{HowardPALSCWCTC19}, the observation $o_k$ is transformed into hidden features $h_k$ via $\operatorname{Enc}\colon o_k \mapsto h_k$. A gated recurrent unit (GRU) \cite{ChoMGBBSB14} cell is used to predict $L$ future waypoints $(x_{k+l})_{l \in \llbracket1, L\rrbracket}$. At each time step $k+l$, the GRU cell updates its hidden state to $h_{k+l+1} = \text{GRU}(h_{k+l}, [x_{k+l}, x_{\text{target}}, x_{\text{target}} - x_{k+l}])$, using position $x_{k+l}$, target point $x_\text{target}$, and their difference as inputs. The hidden state is then decoded by $
      \operatorname{Dec}(\cdot)$ to predict the displacement towards the next waypoint $x_{k+l+1} = x_{k+l} + \allowbreak \operatorname{Dec}(h_{k+l+1})$.  

    \item \textbf{Controller}: We follow \cite{hanselmann2022king} by using PID controller to estimate ego actions from the predicted waypoints $(x_{k+l})_{l\in\llbracket1, L\rrbracket}\mapsto a_k$.
\end{enumerate*}
The planner (2) is trained to learn to map from $(o_k, x_\text{target})$ to future waypoints $(x_{k+l})_{l \in \llbracket1, L\rrbracket}$. Training is done via imitation learning using supervised input-output examples of this mapping.

\paragraph{Action Estimation for Background Agents.}  
As mentioned in \cref{sec:king}, in real-world scenarios, only a series of waypoint coordinates $(x_k)_k$ is known and the actions must be estimated. We use an invertible bicycle model to estimate actions using local derivatives, accurately recovering physically plausible trajectories. Note that this approach may produce noisy actions due to data noise; we leave addressing this issue to future work.

\paragraph{Simulation and Optimization.}
We simulate the scenarios using the AIM-BEV planner for the ego vehicle and estimated actions for the background agents. By leveraging Waymax, we can efficiently compute the exact gradient of the cost \wrt the actions throughout the entire trajectory. This is in contrast to the approach used in \cite{hanselmann2022king}, which uses an approximate gradient to reduce calculation overhead by stopping gradients at specific moments. For both \method{} and \cite{hanselmann2022king}, we use the Adam optimizer with a learning rate of $10^{-3}$ and a maximum of 500 optimization iterations. The optimization process is unsuccessful if it fails to generate a collision.

\subsection{Effectiveness of \method{}}
\label{sec:exp:effectiveness}
\paragraph{Evaluation Protocol.} We conducted our experiments using 200 scenarios extracted from the validation set of WOMD \cite{womd}. First, we validated the ego agent using the original scenarios. Then, we generated safety --- critical scenarios with \method{} and KING and compared their effectiveness both qualitatively --- illustrated with examples --- and quantitatively --- based on the number of scenarios successfully generated and their impact on the ego agent's performance.

\paragraph{Ego Agent Validation.} Since our objective is to enhance an ego agent, we start with one that can drive correctly in regular scenarios. The following scores of the ego planner are reported for these scenarios:
\begin{itemize}[\textbullet]
    \item \textbf{Road Completion} measures the average percentage of the route that the planner completes by the horizon of each scenario or before collision.
    \item \textbf{Infraction Score} is an accumulated penalty when the ego agent collides with another agent, a cyclist, a pedestrian, or ran out of the road boundary. We implemented the score for WOMD with the same penalty rate as \cite{hanselmann2022king}. 
    \item \textbf{Driving score} is the product of the two previous scores. 
    \item \textbf{Collision rate} is the rate of collisions of the ego agent.
\end{itemize}

Our trained agent can reach given target points with an average route completion rate of 88.05\%, indicating acceptable performance. However, despite the high route completion rate, the collision rate is 25.50\%. 

\begin{figure}[ht]
    \centering
    \subcaptionbox{\method{} removes diverging behavior \label{fig:comparison-a}}{
    \begin{minipage}{0.225\linewidth}
\includegraphics[trim={0.648cm 0.654cm 0 0},clip,width=\linewidth]{imgs/king_scenario_00049_60.pdf}\newline
    \includegraphics[trim={0.648cm 0.654cm 0 0},clip,width=\linewidth]{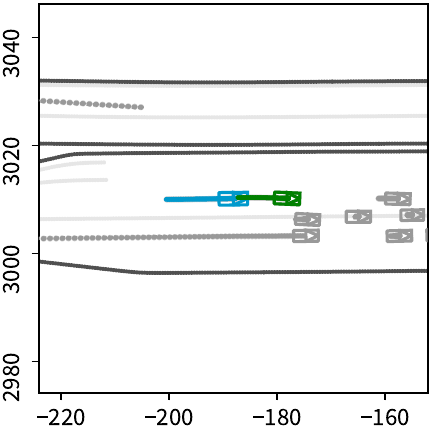}
    \end{minipage}}
    \subcaptionbox{\method{} avoids rear-end\label{fig:comparison-b}}{
    \begin{minipage}{0.225\linewidth}
        \includegraphics[trim={0.648cm 0.654cm 0 0},clip,width=\linewidth]{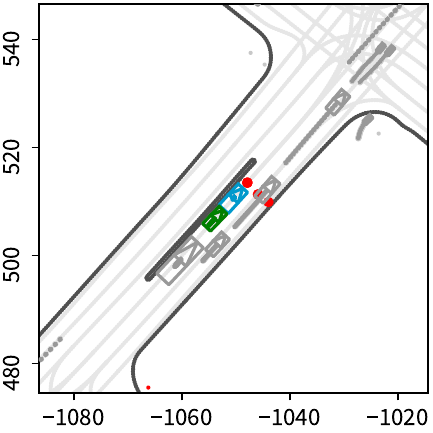}\newline
        \includegraphics[trim={0.648cm 0.654cm 0 0},clip,width=\linewidth]{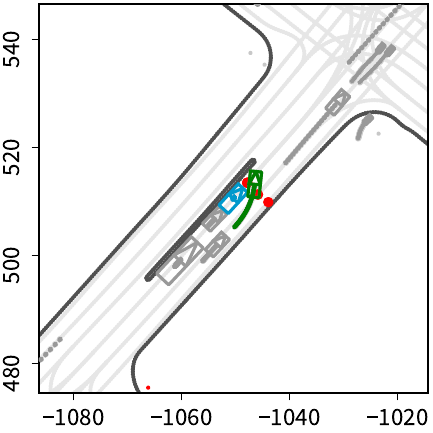}
    \end{minipage}
    }
    \subcaptionbox{\method{} ignores static vehicles\label{fig:comparison-c}}{
    \begin{minipage}{0.225\linewidth}
    \centering
        \includegraphics[trim={0.648cm 0.654cm 0 0},clip,width=\linewidth]{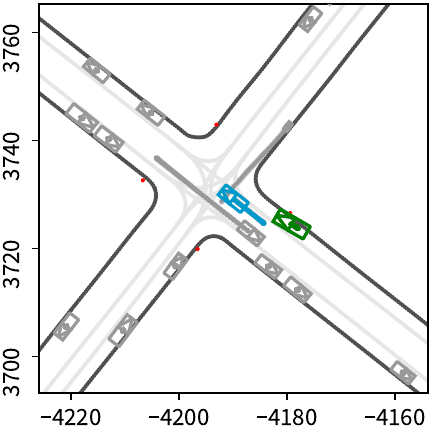}\newline
        \includegraphics[trim={0.648cm 0.654cm 0 0},clip,width=\linewidth]{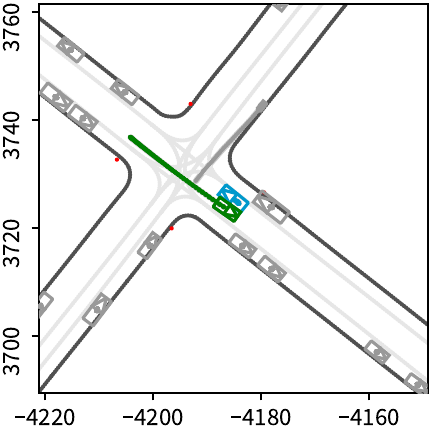}
    \end{minipage}
    }
    \subcaptionbox{\method{} ignores static vehicles and avoids rear-end\label{fig:comparison-d}}{
    \begin{minipage}{0.225\linewidth}
        \includegraphics[trim={0.648cm 0.654cm 0 0},clip,width=\linewidth]{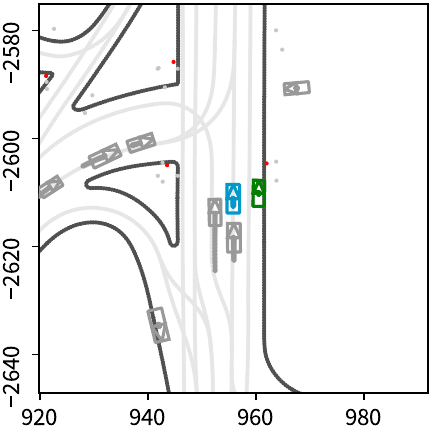}\newline
        \includegraphics[trim={0.648cm 0.654cm 0 0},clip,width=\linewidth]{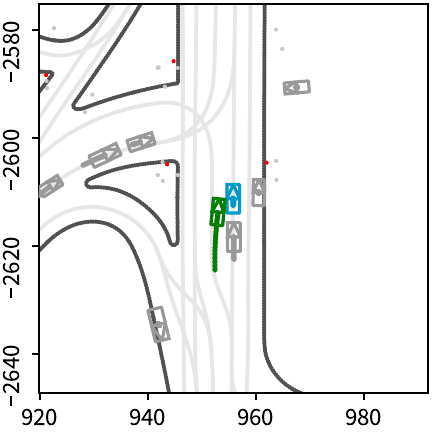}
    \end{minipage}
    }
    
    \caption{\textbf{Qualitative Comparison of Generated Scenarios.} The first row shows the solutions provided by KING \cite{hanselmann2022king}, while the second row displays \method{}’ results. The final state of each scenario is shown along with the trajectory history in dotted line. The ego agent (\textcolor{cyan}{blue}) is at the center, and the adversary agent (\textcolor{ForestGreen}{green}) represents the last optimized adversary. Our method provides non-divergent trajectories and choose better adversary to create more interesting cases.}
    \label{fig:qualitative}
\end{figure}

\paragraph{Qualitative Results.} In \cref{fig:qualitative}, we present a qualitative comparison of \method{} with KING. In \cref{fig:comparison-a}, \cite{hanselmann2022king} makes the optimized adversary run into another background vehicle, despite regularization, while \method{} can steadily decelerate the adversary without creating diverging trajectories. In \cref{fig:comparison-b}, \method{} selects a background agent that provokes a merging collision, in contrast to \cite{hanselmann2022king}, which results in a less interesting rear-end collision. In \cref{fig:comparison-c}, \method{} removes the static agent and creates a head-on collision with the oncoming agent, whereas KING focuses on a stopped vehicle and attempts to start it from the stopping state. In \cref{fig:comparison-d}, \method{} ignores the stopping agent on the right and the other agent behind with risk of rear-end, and modifies marginally the agent on the left to create a lane merging collision from the right side. 
These qualitative results indicate better stability in adversary trajectory generation and improved adversary selection in \method{} compared to \cite{hanselmann2022king}.

\begin{table}[t]
    \centering
    \caption{\textbf{Comparison of Collision Generation Effectiveness for \method{} and \cite{hanselmann2022king}.} The table presents the ego agent's performance scores across 200 original WOMD validation scenarios, as well as those modified by \method{} and \cite{hanselmann2022king}, alongside their generation success rates. ``$\uparrow$ for generation'' indicates that a higher value is better for collision generation, vice versa. The metrics are defined in the beginning of \cref{sec:exp:effectiveness}.}
    \label{tab:aim-bev-val-test}
    \begin{tabular}{llccc}
    \toprule
     \multicolumn{2}{l}{WOMD Val Scenarios}  & Original & KING \cite{hanselmann2022king} & \method{}
     \\
    \midrule
    \multicolumn{2}{l}{Generation Success Rate ($\uparrow$ for generation)} & n/a & 48.99 & \textbf{60.40} \\
\midrule
 \multirow{4}{*}{\rotatebox[origin=c]{90}{\shortstack[c]{\tiny \textit{Ego Agent} \\ \tiny \textit{Performance}}}} & Route Completion ($\uparrow$ for ego, $\downarrow$ for generation)  & 88.05 & 73.68 & 71.06 \\
   & Infraction Score ($\uparrow$ for ego, $\downarrow$ for generation) & 0.82 & 0.72 & 0.69 \\
   & Driving Score ($\uparrow$ for ego, $\downarrow$ for generation) & 74.20 & 56.19 & 50.78 \\
   & Collision Rate ($\downarrow$ for ego, $\uparrow$ for generation) & 25.50 & 59.00 & 69.50 \\
    \bottomrule
    \end{tabular}
\end{table}

\paragraph{Quantitative Results.}
To demonstrate the effectiveness of collision generation, we report \textbf{Generation Success Rate}, the percentage of collisions that corresponding method have successfully created among the previously collision-free scenarios. In \cref{tab:aim-bev-val-test}, we present the results comparing scenarios modified by \method{} with those generated by the original KING. 

We observe a significant drop in the collision generation success rate for KING, decreasing from approximately 80\% with synthetic scenarios in the original work to 49\% with WOMD. This decline highlights the challenges of applying this method to real-world settings. In contrast, \method{} improves the success rate to 60\% with WOMD, representing a 10 percentage point increase.

Additionally, both \method{} and KING generate collisions that negatively impact the ego vehicle's driving performance. \method{} tends to penalize the ego agent more, likely due to the higher number of collisions created.

\subsection{Notes for Implementation}
Implementing our approach with Waymax and JAX required reworking most components from scratch to ensure efficient optimization. 

We leveraged just-in-time (JIT) compilation and function vectorization to parallelize and accelerate most parts of the calculation for optimal performance on GPUs. New functions were implemented for rasterizing observations directly on GPUs, accelerating the ego agent's planning and scenario simulation.

Due to the lack of some information in the WOMD dataset, cost terms were adapted based on available information. For example, no drivable area map was directly provided in WOMD, so we rebuilt it using road boundary information. For the same reason, some evaluation metrics were adapted, such as the out-of-road penalty score, which requires access to the drivable area map.

Additionally, we smoothly integrated a pretrained neural network with PyTorch into our JAX pipeline using \texttt{torch2jax} \cite{torch2jax}, avoiding potential performance issues caused by network architecture re-implementation and weight transfer. 

\section{Conclusion}

In this work, we propose \method{}, which enhances the stability of kinematic model-based trajectory optimization methods. We address issues discovered after scaling the approach from \cite{hanselmann2022king} for use in real-world settings. Our method identifies and mitigates issues that cause instabilities and generate unactionable scenarios. It generates more meaningful scenarios in complex settings and provides a robust and flexible technical foundation for developing efficient solutions for safety-critical scenario generation.
In future work, it would be valuable to investigate how well the generated corner-cases, which are currently optimized to a specific ego agent's planning algorithm, generalize to different driving agents. Additionally, exploring whether it is possible to design corner-cases that universally cause planners to fail would be an interesting avenue to pursue.

\section*{Acknowledgements}
This work was supported by the ANR grant MultiTrans (ANR-21-CE23-0032). We thank the anonymous reviewers for their remarks. This work was made using the
Waymax Licensed Materials, provided by Waymo LLC under the Waymax License Agreement for Non-Commercial Use, available at \url{https://github.com/waymo-research/waymax/blob/main/LICENSE},
and your access and use of the Waymax Licensed Materials are governed by the terms and conditions contained therein.

\bibliographystyle{splncs04}
\bibliography{bib}

\end{document}

%% file: math_commands.tex

\crefname{equation}{eq.}{eqs.}
\Crefname{equation}{Eq.}{Eqs.}
\crefname{assumption}{assumption}{assumptions}
\Crefname{assumption}{Assumption}{Assumptions}
\crefname{definition}{definition}{definitions}
\Crefname{definition}{Definition}{Definitions}
\crefname{prop}{proposition}{propositions}
\Crefname{prop}{Proposition}{Propositions}
\crefname{theorem}{theorem}{theorems}
\Crefname{theorem}{Theorem}{Theorems}
\crefname{example}{example}{examples}
\Crefname{example}{Example}{Examples}
\crefname{page}{p.}{pp.}

\let\originalleft\left
\let\originalright\right
\renewcommand{\left}{\mathopen{}\mathclose\bgroup\originalleft}
\renewcommand{\right}{\aftergroup\egroup\originalright}


\def\1{\bm{1}}











\DeclareMathAlphabet{\mathsfit}{\encodingdefault}{\sfdefault}{m}{sl}
\SetMathAlphabet{\mathsfit}{bold}{\encodingdefault}{\sfdefault}{bx}{n}


\def\gC{{\mathcal{C}}}
\def\gD{{\mathcal{D}}}

\def\gK{{\mathcal{K}}}

\def\gQ{{\mathcal{Q}}}

\def\gS{{\mathcal{S}}}










\newcommand{\R}{\mathbb{R}}



\DeclareMathOperator*{\argmin}{arg\,min}

\newcommand{\eg}{e.g.\@\xspace}
\newcommand{\sut}{s.t.\@\xspace}
\newcommand{\ie}{i.e.\@\xspace}

\newcommand{\wrt}{w.r.t.\@\xspace}


\let\vec\boldsymbol


\DeclarePairedDelimiterX{\midx}[2]{(}{)}{#1\;\delimsize\vert\;#2}
\DeclarePairedDelimiterX{\midbracesx}[2]{\{}{\}}{#1\;\delimsize\vert\;#2}
\DeclarePairedDelimiterX{\parallelx}[2]{(}{)}{#1\;\delimsize\|\;#2}



\definecolor{cpink}{HTML}{FCCDE5}
\definecolor{cred}{HTML}{FFC2BA}
\definecolor{cyellow}{HTML}{FFFFB3}
\definecolor{cblue}{HTML}{B9DEFF}
\definecolor{cgreen}{HTML}{D7F3E7}
\definecolor{cneutral}{HTML}{CFCFCF}


%% file: figs/teaser.tex
\begin{figure}[htbp]
    \centering
    \subcaptionbox{Original \label{fig:teaser-original}}{\frame{\includegraphics[trim={0.68cm 2.2cm 0.8cm 1.1cm},clip,width=0.3\linewidth]{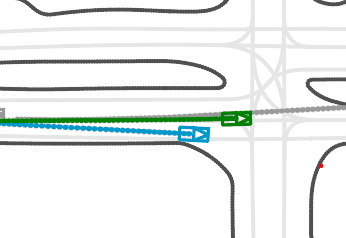}}}\hspace*{\fill}
    \subcaptionbox{Without constraints \cite{hanselmann2022king} \label{fig:teaser-king}}{\frame{\includegraphics[trim={0.68cm 2.2cm 0.8cm 1.1cm},clip,width=0.3\linewidth]{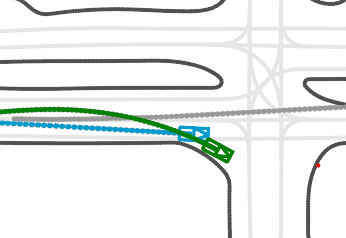}}}\hspace*{\fill}
    \subcaptionbox{Ours: \method{} \label{fig:teaser-ours}}{
    \frame{\includegraphics[trim={0.68cm 2.2cm 0.8cm 1.1cm},clip,width=0.3\linewidth]{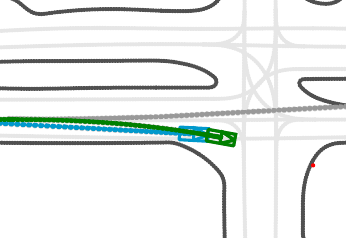}}}
    \caption{
    \textbf{Importance of Constraints in \textcolor{cyan}{Ego}-\textcolor{ForestGreen}{Adversary} Collision Scenarios.} The figures depict the final state with past trajectories shown as dotted lines. The original collision-free scenario \subref{fig:teaser-original} is modified by an unconstrained method \cite{hanselmann2022king} \subref{fig:teaser-king} and our \method{} \subref{fig:teaser-ours} with proposed constraints. The \textcolor{cyan}{ego} vehicle is at the center, and the \textcolor{ForestGreen}{adversary} is in front. In \subref{fig:teaser-king}, the unconstrained adversary takes an unrealistic swinging turn (first to the left then to the right), leading to a partially diverging trajectory that does not reflect plausible real-world driving behavior. In contrast, our \method{} \subref{fig:teaser-ours} ensures the adversary decelerates smoothly while gradually approaching the ego, maintaining a realistic and natural trajectory.
    }
\end{figure}